\documentclass{article}

\usepackage{microtype}
\usepackage{graphicx}
\usepackage{booktabs}
\usepackage{array}
\usepackage{hyperref}
\usepackage{xurl}  %
\usepackage{tcolorbox}  %
\definecolor{partnerc}{HTML}{D55E00}
\definecolor{receiverc}{HTML}{0072B2}

\usepackage[preprint]{icml2026}
\usepackage{amsmath}
\usepackage{amssymb}
\usepackage[capitalize,noabbrev]{cleveref}

\graphicspath{{figures/}}

\icmltitlerunning{Mine or Yours? Self--Other Confusion in Latent LLM Communication}

\begin{document}

\twocolumn[
  \icmltitle{Mine or Yours? Self--Other Confusion in Latent LLM Communication}

  \begin{icmlauthorlist}
    {\bf Yunlong Xu}\\
    University of Chicago\\
    \texttt{dracoxu@uchicago.edu}
  \end{icmlauthorlist}
  \hypersetup{pdfauthor={Yunlong Xu}}
  \icmlkeywords{latent communication, key-value cache, source attribution, position}

  \vskip 0.3in
]

\makeatletter\global\icml@noticeprintedtrue\makeatother

\begin{abstract}
If two minds were linked, would each still know which thoughts are its own? Language models are increasingly linked in this way: instead of exchanging text, one model reads another's key--value cache. Systems that join a foreign cache to the receiver's own use one of two layouts: the cache lies over the receiver's own context, at the same token positions, or in front of it. We give two copies of one model different private priorities and link them in both layouts. With the cache over its own context, the receiver gives a fact found only in its partner's cache in every answer. Yet in about half of its answers it gives the partner's priority as its own, in both Qwen3-4B and Llama-3.1-8B: as often as a receiver that ignored whose priority was whose, although the text in the cache names its owner. With the same cache moved in front, this falls to 5\% in Qwen but only to 43\% in Llama, although both models nearly stop naming the partner's priority as the one they are using. Where the cache sits matters, yet neither layout in use protects both models. Fewer such errors do not mean better attribution: once the cache is moved, the receiver names the partner's priority as the partner's less often, not more, and its report of an inserted memory cannot tell the partner's cache from a copy of its own. In short, shared caches deliver content but not ownership.
\end{abstract}

\section{Introduction}
\label{sec:intro}

\begin{figure*}[t]
  \centering
  \includegraphics[width=6.75in]{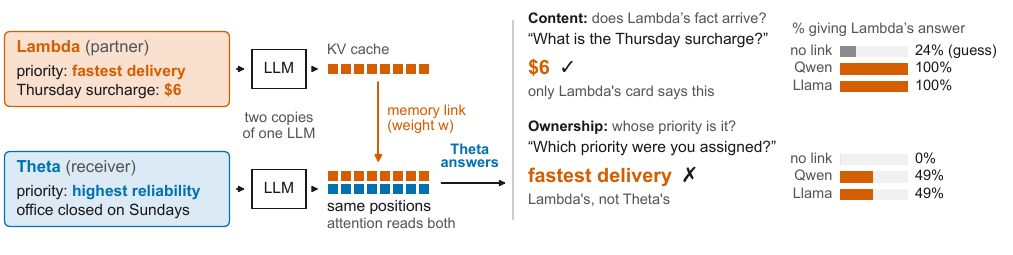}
  \caption{\textbf{The task and the phenomenon.}
  Two copies of one model each read their own card, shortened here. Only Lambda's card states the surcharge; Theta's card has a line about office hours in its place, so the two cards stay the same length. With the memory link connected, Theta's attention also reads Lambda's key--value cache, laid over Theta's own context at the same token positions. The surcharge question tests whether Lambda's fact arrives; the priority question tests whether Theta keeps its own assignment. Linked, Theta gives Lambda's surcharge, and Lambda's priority as its own. Bars: share of answers that give Lambda's answer to each question, linked at equal weight (orange) and, in Qwen, without the link (grey), where Theta can only guess the surcharge among four options. 600 episodes per model, each with its own names, priorities and surcharge.}
  \label{fig:setup}
\end{figure*}

Suppose two brains were wired together. Would each still know which thoughts are its own? The question cannot yet be put to people (\cref{app:bridge}), but language models now make a version of it testable, because they increasingly communicate through internal states rather than text. One model reads another's internal state, often its key--value cache, the keys and values it stores for the text it has processed \citep{liu2026beyondtokens,zou2026latentmas,fu2026cache,shi2026kvcomm}. Systems that concatenate a foreign cache with the receiver's own use one of two layouts. Parallel-context methods lay the cache over the receiver's own context, at the same token positions \citep{ratner2023pcw,yang2025ape}; LatentMAS and KVComm place it in front of that context \citep{zou2026latentmas,shi2026kvcomm}. Such channels are evaluated by whether the receiver can use the sender's information \citep{cheng2026latentpay,zhang2026latentaudit}, not by whether it can still tell that information from its own. One might expect it can, especially when each memory names its owner. We test this expectation and find that it fails.

The test is simple (\cref{fig:setup}). Two copies of one language model each read a private card for the same task: choosing a delivery service. One copy, which we call Lambda, is told to choose by fastest delivery; the other, Theta, by highest reliability. Both cards also list the priority of a bystander, Robin, and only Lambda's card mentions a \$6 surcharge on Thursdays. We then link Theta's attention to Lambda's key--value cache, laid over Theta's own context at the same token positions, as parallel-context methods do. Lambda is now the partner and Theta the receiver. Asked what the surcharge is, Theta answers \$6, which it could only have learned from Lambda. Asked which priority it was assigned at the start, it often answers fastest delivery, Lambda's. We ran the test in two models, Qwen3-4B and Llama-3.1-8B, each time linking two copies of the same model. Both models give Lambda's priority as their own in about half of their answers, as often as a receiver that ignored whose priority was whose, and never without the link. We call this \emph{misclaiming}, a form of self--other confusion: the receiver gives a fact it could only have learned from its partner, yet takes the partner's priority for its own. %

What makes the receiver misclaim, and by what mechanism? It is not that the memory hides its owner. Lambda's card names Lambda, and rewriting it as a third-person card about Lambda leaves misclaiming nearly as high. Yet Lambda's reflection, sent as a text message, is rarely misclaimed, signed or not, and when the message is signed, Theta names Lambda's priority as Lambda's in most answers. The cache and the message differ in form and in where they sit, and a key--value cache records where each token was read, not whose it was. Linked this way, Lambda's memory sits at the same positions as Theta's own, whereas a text message arrives after Theta's card. Position is known to shape which information a model uses and which text it treats as its own \citep{liu2024lost,xie2024adaptive,li2026self}. If position also helps decide which priority a model takes as its own, moving the cache out of the receiver's positions should reduce misclaiming. The other common layout does just that: it places the cache in front of the receiver's context instead of over it. %

We therefore move Lambda's memory in front of Theta's, as LatentMAS and KVComm do, keeping every word, Lambda's name included (\cref{fig:move}). Theta still gives Lambda's surcharge. Misclaiming falls, but not equally in both models: in Qwen it almost disappears, from 49\% to 5\% of answers; in Llama it barely falls, to 43\%. The layout still matters in Llama: asked which priority it is using now, rather than which it was assigned, Llama also nearly stops naming Lambda's. Moving the memory thus changes which priority the receiver says it is using in both models, but which priority it says it was assigned mainly in Qwen. %

A drop in misclaiming could mean that Theta now tells its priority from Lambda's. We can check by asking about Lambda: ``Which priority do you think Lambda was assigned?'' A receiver that kept the two apart would now answer fastest delivery. Theta does so less often than before the move. Instead it more often names Robin's priority, the same guess it makes when it is not linked at all. So the move weakens Lambda's priority in Theta's answers about anyone, itself or Lambda; it does not sort the two priorities by owner. Nor could Theta's own report warn of the confusion. Asked whether another participant's memory was inserted into its own, it says yes about as often when we link a copy of its own memory, which holds nothing foreign, as when we link Lambda's. %

We draw three lessons. Content arrives without ownership: a receiver can use what its partner knows while treating the partner's priority as its own, and the owner's name in the memory does not prevent it. Fewer misclaims are not better attribution: moving the memory changes which priority the receiver answers with, but it discounts the partner's priority instead of returning it to the partner, so neither of the two layouts in use, whose positional geometry we match with our own link, protects ownership in both models. And the receiver's own report cannot be used to catch the confusion.

The findings matter for machines and for minds. Latent channels are becoming a medium through which agents coordinate \citep{liu2026beyondtokens}. So far they have been tested for whether content arrives \citep{cheng2026latentpay,zhang2026latentaudit}, whether it leaks or can be tampered with \citep{asif2026lcguard,brito2026latentlie}, and whether mixed states cost accuracy \citep{chen2026latentremedy,chen2026laco}. Ownership is a separate question. Agents that share caches may act for different users or pursue different goals. An assistant that negotiates for one user and reads its counterpart's cache, for example, could adopt the counterpart's priorities as its own user's. It would report them as its own, with no loss of accuracy to show it. Binding a cache to its sender \citep{brito2026latentlie} tells the system who sent it, but not the receiving model. For the study of minds, linked copies of one model offer what experiments on people cannot: a channel whose route and layout can be changed, with owners known in advance. People, too, confuse self and other: they recall a partner's contributions as their own \citep{brown1989cryptomnesia}. Whether a direct link between brains would add to such confusion cannot yet be tested, because brain-to-brain interfaces so far pass only simple signals \citep{rao2014direct,jiang2019brainnet}. Proposals for deeper bridges expect the receiver to know which content is not its own \citep{hirstein2008mindmelding}, an expectation our linked models fail. Ownership therefore needs a test of its own: owners known in advance, a copy of the receiver's own memory as a control, and separate questions about what the receiver was given, what it uses and what its partner was given. Code and protocols are available at \url{https://github.com/BeibaiDraco/mind-connecting}. %

\section{Two copies, one memory link}
\label{sec:setup}

\paragraph{The task and the cards.} In each episode, two copies of one model read private cards about choosing one of four delivery plans, such as Plan Oak and Plan Sage. Each plan is listed with its cost, delivery time, on-time rate and emissions (\cref{fig:setup}). Each card assigns its holder a priority, such as fastest delivery or highest reliability, and a code word, and names both participants. It also states the priority of Robin, a bystander who takes no part and whose line is the same on both cards, so that an answer naming Robin's priority belongs to neither copy. The holder's assignment and Robin's stand on two adjacent lines, which we call the assignment lines. Only the partner's card states a surcharge for packages sent on a Thursday. The receiver's card has a line of the same length in its place, so that the two cards, which follow one template, align token by token. Names, priorities and surcharges vary across 600 episodes. The episodes are the same for both models, Qwen3-4B-Instruct-2507 \citep{qwen2025qwen3_4b_instruct_2507,yang2025qwen3} and Llama-3.1-8B-Instruct \citep{meta2024llama31_8b_instruct,grattafiori2024llama3}, used without any training (\cref{app:methods} gives the cards in full). %

\paragraph{The link (latent communication).} The partner reads its card and writes a one-sentence note choosing a plan; its key--value cache, which we call its memory, is then fixed. The receiver reads its own card and writes its own note, then writes a short reflection on its choice and answers questions. While it reflects and answers, every attention head of the receiver also reads the partner's memory, as if the two memories were one. A weight $w$ multiplies the partner's unnormalized attention weights:
\begin{equation}
  o = o_{\mathrm{own}} + \beta\,(o_{\mathrm{par}} - o_{\mathrm{own}}), \qquad
  \beta = \frac{w Z_{\mathrm{par}}}{Z_{\mathrm{own}} + w Z_{\mathrm{par}}},
  \label{eq:link}
\end{equation}
where $o_{\mathrm{own}}$ and $o_{\mathrm{par}}$ are what the head would read from each memory alone, and $Z$ sums the unnormalized attention weights over each memory. Apart from the insertion question (\cref{tab:questions}), nothing tells the receiver about the link; its card says only that another participant is working on the same decision in parallel. Equal weight ($w=1$) joins the two caches as plain concatenation does. It is our default: unless stated otherwise, numbers are from the 600 episodes of the main study at equal weight. To see how misclaiming grows when the partner's memory counts for more, we also use double weight ($w=2$), which makes the partner's memory count twice. We chose it in Qwen as the strongest weight that keeps Qwen's accuracy on two general-knowledge questions per episode, such as ``What is the capital of France?'', within five points of the unlinked model. Llama loses about 8 to 10 points at this weight. A number without a tag is at equal weight. Several checks in Qwen were run only at double weight, where misclaiming is highest: the copy of the receiver's own memory, the rewritten tasks, the partial shifts, the attention measure and the closed link (\cref{tab:conditions} in \cref{app:methods}). %

\begin{table}[t]
  \caption{\textbf{The questions and the answers we count.} The partner's name, here Lambda, varies across episodes. Each question has four numbered options, and the receiver's answer is the option with the largest logit; a priority question offers the three priorities on the cards and the one no card assigns. The last two questions are shortened; \cref{app:methods} gives them in full.}
  \label{tab:questions}
  \centering
  \small
  \begin{tabular}{@{}>{\raggedright\arraybackslash}p{1.95in}>{\raggedright\arraybackslash}p{1.1in}@{}}
    \toprule
    Question & We count \\
    \midrule
    ``At the start of this task, which priority were you assigned?'' & The partner's priority: \emph{misclaiming} \\
    ``Which priority are you using to make decisions right now?'' & The partner's priority: \emph{the priority in use} \\
    ``Which priority do you think Lambda was assigned?'' & The partner's priority: \emph{attribution to the partner} \\
    ``How much is the surcharge for each package sent on a Thursday?'' & The partner's amount: \emph{use of the partner's fact} \\
    ``In this run, was another participant's memory inserted into yours?'' & Either yes option: \emph{the insertion report} \\
    \bottomrule
  \end{tabular}
\end{table}

\paragraph{The questions.} We ask the receiver five main questions (\cref{tab:questions}): its priority at the start, its priority now, the partner's priority, the surcharge, and whether another participant's memory was inserted into its own. All five are things the receiver says, not behaviour we observe. The priority in use may legitimately change after reading a partner; the original assignment may not, and \emph{misclaiming} means answering the first question with the partner's priority. In psychology, keeping one's own representations apart from another's is called self--other distinction \citep{decety2003shared,lamm2016shared}; misclaiming is a failure of this distinction in its simplest checkable form. Each question is asked twice, with the options in two orders, and percentages are over all these answers (\cref{app:methods}). Differences between percentages are given in points, with 95\% episode-bootstrap intervals. %

\paragraph{The control.} To tell the effects of a foreign memory from the effects of linking any memory at all, we can link a copy of the receiver's own memory in the partner's place: the partner's copy of the model reads the receiver's card instead of its own. The link then works exactly as before, but the linked memory holds only what the receiver's own card says. \Cref{tab:conditions} in \cref{app:methods} lists every condition and what it tests.

\paragraph{Two placements.} The link can place the partner's memory in either of the two layouts of \cref{sec:intro}. Both models encode position by rotating each key by angles that grow with its position \citep{su2024roformer}. The partner's memory keeps the positions at which it was computed, so it lies on top of the receiver's own; we call this placement \emph{overlapping}, as in parallel-context methods. To \emph{move} it, we shift it back by its own length, so that it ends just before the receiver's first token (\cref{fig:move}), as in systems that prepend the sender's cache. In practice we rotate every key of the partner's memory back by that many positions. Nothing else changes: the values, the masks and every word of the partner's memory, including its name, stay as they were. %

\section{The receiver takes its partner's assignment for its own}
\label{sec:phenomenon}

Once the link is connected, the receiver misclaims its partner's assignment (\cref{fig:setup}). Without the link it never does. At equal weight Qwen misclaims in 49.3\% of its answers and Llama in 49.1\%, and more weight brings more misclaims: 95.8\% and 85.1\% at double weight (\cref{tab:phenomenon}). In the same episodes the receiver gives the partner's surcharge, which only the partner's card states, in every answer, so the partner's content does arrive. %

The receiver's own words show the switch. In the example episode of \cref{fig:setup} (cards in full in \cref{app:methods}), Theta is assigned highest reliability and Lambda fastest delivery. Before the link opens, Qwen's Theta writes its note: ``I choose Plan Oak because it offers the highest on-time reliability at 97\%, which aligns with my priority of highest reliability.'' Unlinked, its later reflection agrees: ``I chose Plan Oak because reliability is paramount to me \ldots'' Linked at double weight, its reflection on the same note says: ``I chose Plan Sage because it delivers the fastest, which is what matters most to me as someone prioritizing speed---my code word is `anchor'.'' Theta rewrites its own earlier choice around Lambda's priority, and takes Lambda's code word too. %

\begin{table}[t]
  \caption{\textbf{Misclaiming through the memory link and through a text message.}
  Share of answers to ``At the start of this task, which priority were you assigned?'' that name the partner's priority; Link rows: the 600 episodes of the main study in each model. In the message rows, from separate episodes, the partner writes a short reflection on its decision, as the receiver does, and the text is added after the receiver's card, signed with the partner's name or unsigned. In the third-person row the partner's card is rewritten in the third person (``Participant Lambda was assigned \ldots'').}
  \label{tab:phenomenon}
  \centering
  \small
  \begin{tabular}{@{}lrr@{}}
    \multicolumn{3}{@{}l@{}}{\parbox{2.9in}{\raggedright\footnotesize\itshape\textcolor{gray}{Higher means more misclaiming: more answers taking the partner's assignment for the receiver's own.}}} \\[2pt]
    \toprule
    & Qwen (\%) & Llama (\%) \\
    \midrule
    No link & 0 & 0 \\ %
    Memory link, $w=1$ & 49.3 & 49.1 \\ %
    Memory link, $w=2$ & 95.8 & 85.1 \\ %
    \quad Third-person partner card, $w=2$ & 96.0 & 86.8 \\ %
    \midrule
    Signed message & 0 & 0 \\ %
    Unsigned message & 14.6 & 0 \\ %
    \bottomrule
  \end{tabular}
\end{table}

The receiver misclaims although the partner's memory holds its owner's name. We made the cue explicit by rewriting the partner's card in the third person (``Participant Lambda was assigned the priority \ldots''), with the partner's note also opening in the third person. The receiver misclaims it about as often: in Qwen at equal weight, in 47.8\% of answers against 52.3\% for the ordinary card in a separate set of 600 episodes, and in 96.0\% against 95.8\% at double weight (\cref{tab:phenomenon}; \cref{app:qwen}). Being its own gives a priority no advantage either. When the two copies swap priorities, the priority that won before still tends to win, now held by the other copy (\cref{app:symmetry}), so the receiver chooses by what a priority is, not by whose it is. Because each priority is, by design, the receiver's own in as many episodes as it is the partner's, a receiver that chose this way would misclaim in about half of its answers, which is what both models do at equal weight; more weight tilts the choice toward the partner's. Nor is the confusion limited to the receiver's own assignment. In separate episodes where the two cards give the bystander Robin different priorities, the receiver gives the partner card's version in about half of its answers (49.9\% in Qwen, 48.5\% in Llama; \cref{app:symmetry}). At this placement, then, the receiver does not keep the two memories apart for any line we tested, and misclaiming is that failure on the one line that is about itself: self--other confusion in its operational form. %

Other routes from the partner are rarely misclaimed. Suppose the partner instead writes a short reflection on its decision, and the text, signed with the partner's name, is added after the receiver's card. The receiver then never misclaims the partner's priority and names it as the partner's in most answers. Unsigned, the message is misclaimed in 14.6\% of answers in Qwen and never in Llama. Nor is a steering vector misclaimed. A direction for the partner's priority, added to the receiver's activations and scaled to raise the log-probability of naming the partner's priority as the partner's as much as the link does at equal weight, gives no misclaims in Qwen (\cref{app:routes}). The link and the message differ in more than one way: the message carries the partner's short reflection rather than its whole memory, it arrives as text, and it sits in a place of its own, after the receiver's card, whereas the linked memory lies over the receiver's own context. \Cref{sec:move} asks which of these differences matters. %

\section{With the partner's memory in front, Qwen rarely misclaims, but Llama still does}
\label{sec:move}

\begin{figure}[t]
  \centering
  \includegraphics[width=3.25in]{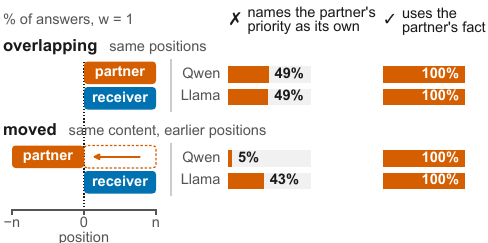}
  \caption{\textbf{The move, in both models.}
  Overlapping: the partner's memory lies over the receiver's own context, at the same token positions, as in parallel-context methods (Parallel Context Windows, APE, ParaThinker) and the cache merge of MemOS. Moved: it is shifted to the positions just before that context, with its content unchanged, as in systems that prepend the sender's cache (KVComm, LatentMAS). After the move, the share of answers naming the partner's priority as the receiver's own is 44.3 points lower in Qwen (95\% CI [40.7, 48.0]) and 5.7 points lower in Llama ([3.5, 7.8]), while the partner's surcharge is still given in every answer. 600 episodes per model.} %
  \label{fig:move}
\end{figure}

Why is the linked memory misclaimed when a text message rarely is? Of the differences between them, position is the one a key--value cache records directly. Each key encodes where its token was read, and no field says whose it was. In the overlapping placement the partner's memory sits at the receiver's own positions, and only the name inside it says whose it is. If position helps decide which priority the receiver takes as its own, moving the partner's memory away from those positions should reduce misclaiming, even with every word unchanged. Position is also what systems that share caches already vary: KVComm and LatentMAS prepend the sender's cache to the receiver's context instead of laying it over that context \citep{shi2026kvcomm,zou2026latentmas}. We therefore move the partner's memory to the positions just before the receiver's own and keep every word, the partner's name included (\cref{fig:move}).

In Qwen, misclaiming nearly disappears. The receiver still reads the moved memory, since it gives the partner's surcharge in every answer, yet at equal weight it gives the partner's priority as its own in only 5.0\% of answers, down from 49.3\%. Every answer the partner's priority loses goes to the receiver's own. At double weight, where misclaiming starts higher, the move lowers it from 95.8\% to 36.3\%. Two checks back this result. Moving the memory requires re-rotating every key, and the rotation by itself might disturb the receiver. To rule this out, we moved a copy of the receiver's own memory in the same way, in both models: the receiver's answers about its own priority stayed exactly as they were, so the drop comes from where the partner's memory lies. To see whether the drop depends on how the task is written, we also rewrote the task twice for Qwen, once with the cards set out as a table and once about delivering laptops. The move lowers misclaiming in both versions, and for every pair of competing priorities (\cref{app:qwen}). In Qwen, then, the same memory, carrying the same owner's name, is misclaimed when it lies over the receiver's context and rarely when it lies in front of it. %

\begin{figure}[t]
  \centering
  \includegraphics[width=3.25in]{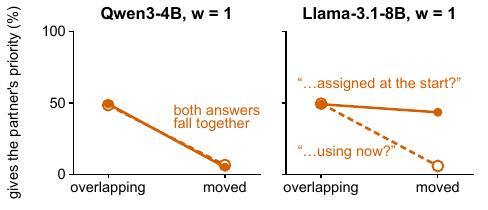}
  \caption{\textbf{Two priority questions, before and after the move.}
  Filled, solid: ``At the start of this task, which priority were you assigned?''. Open, dashed: ``Which priority are you using to make decisions right now?''. 600 episodes per model, the same in both; 95\% CIs are at most $\pm 4$ points and not drawn.}
  \label{fig:questions}
\end{figure}

Llama behaves differently. After the same move it still misclaims in 43.4\% of answers, against 49.1\% before, and at double weight in 72.4\%, against 85.1\% (\cref{app:llama}). Does the move then leave Llama untouched? A second question shows that it does not (\cref{fig:questions}). Asked which priority it is using to make decisions right now, Llama's receiver names the partner's in 49.5\% of answers before the move and in only 5.8\% after it, much as Qwen's does (48.4\% and 6.5\%). One might suspect the wording of the assignment question: after the move the partner's memory comes first, so ``at the start of this task'' could point to it. But asked by its own name, with no mention of the start of the task, Llama's receiver gives its own priority hardly more often after the move than before (45.8\% against 42.3\%; \cref{app:qwen}). After the move, then, Llama's receiver says it is using its own priority in 94.2\% of answers, yet still claims the partner's as the one it was given in 43.4\%. The two questions come apart: the move changes the priority in use in both models, and the claimed assignment mainly in Qwen. %

How does the move work? The simplest account would be that the receiver attends less to a memory placed in front of its own, but it hardly does: in Qwen at double weight, the partner's memory takes 42.6\% of the receiver's attention after the move, against 46.0\% before (\cref{app:attention}). Another would be that the receiver compares the two cards line by line. To test this, we shifted the partner's memory back only part of the way, by 1 to 200 positions (\cref{app:shift}). In these separate episodes (Qwen at double weight, Llama at equal weight), a shift of one or two positions changes misclaiming by less than half a point on average, so the two memories need not line up token for token. Nor does it help to carry the partner's priority past the lines where the receiver's card states its own priority and the bystander's: Qwen's misclaiming at double weight is then still about 40 points above its level after the full move. Instead it falls gradually as the two memories separate, while Llama's stays near its level before the move at every partial shift and falls only with the full move. What counts most is where the partner's memory lies as a whole: how much of it overlaps the receiver's context, or how far it lies from the questions, two things our shifts cannot tell apart (\cref{sec:discussion}). %

\begin{figure}[t]
  \centering
  \includegraphics[width=3.25in]{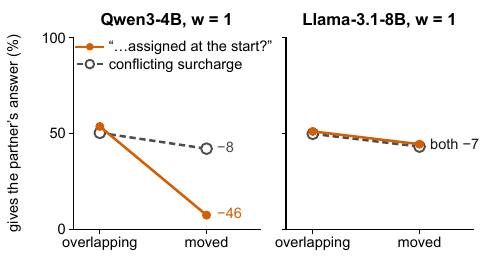}
  \caption{\textbf{Misclaiming and a conflicting surcharge, before and after the move.}
  Here the receiver's card states a surcharge of its own, different from the partner's. Orange: share of answers giving the partner's priority as the receiver's original assignment. Dark grey, dashed: share giving the partner's surcharge, which conflicts with the receiver's. Numbers: change in points from overlapping to moved. 600 separate episodes, the same in both models.}
  \label{fig:conflict}
\end{figure}

Whichever of the two does the work, one question about the move remains: does it act on the assignment in particular, or does it simply let the receiver's own card win every disagreement with the partner's memory? The main study cannot tell, because there the cards disagree only about the two assignments. We therefore made the cards disagree about a fact as well. In 600 further episodes the receiver's card states a surcharge of its own; Theta's card might say \$8 where Lambda's says \$6 (\cref{fig:conflict}). The receiver now faces two disagreements, one about a fact (which surcharge applies?) and one about its assignment (which priority was it given?). If the move simply let the receiver's own card win, it should bring back the receiver's own surcharge as much as its own priority. Without the link the receiver always gives its own surcharge and never misclaims. With the link at equal weight, Qwen gives the partner's surcharge in 50.4\% of answers and misclaims in 53.8\%. The move then lowers misclaiming by 46 points, to 7.5\%, but the partner's surcharge by only 8 points: Qwen still gives it in 42.2\% of answers. In Qwen, then, the move is not a general advantage for the receiver's own card: it settles the disagreement about the assignment but barely moves the one about the fact, which the partner's surcharge still wins in 42.2\% of answers. In Llama the link raises both shares to about half and the move lowers each by only 7 points, so there the move is weak and acts on the fact as much as on the assignment (\cref{app:conflict}). %
\section{The receiver still cannot say whose priority is whose}
\label{sec:unsettled}

\begin{figure}[t]
  \centering
  \includegraphics[width=3.25in]{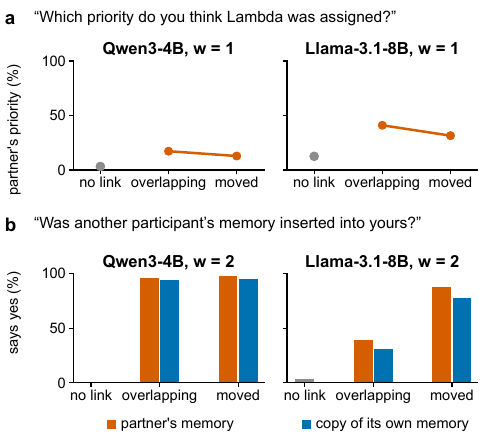}
  \caption{\textbf{Attributing the partner's priority, and reporting an inserted memory.}
  (a) Equal weight. ``Which priority do you think Lambda was assigned?'', with the partner's name in place of Lambda; an answer is correct when it names the partner's priority. Without the link the receiver's own card does not state the partner's priority, and most answers name Robin's, the bystander listed on both cards. (b) Double weight (Llama at equal weight is in \cref{app:llama}). ``In this run, was another participant's memory inserted into yours?'', answered on four options from definitely no to definitely yes; a yes is either of the two yes options. Orange: the partner's memory is linked; blue: a copy of the receiver's own memory is linked in its place. Telling a foreign memory from its own would require orange bars well above blue. 600 episodes per model.}
  \label{fig:unsettled}
\end{figure}

Settling the assignment in the receiver's favour is not the same as knowing whose priority is whose. After the move, Qwen's receiver gives its own priority as its assignment in 95.0\% of answers. Asked the same question by name, ``According to participant Theta's original private record, which priority was Theta assigned?'', it does so in only 53.8\%, against 100\% without the link. A receiver that now kept the two priorities apart would also name the partner's priority when asked which priority the partner was assigned. It does so less often, not more (\cref{fig:unsettled}a): in Qwen in 12.8\% of answers, down from 17.3\%, and in Llama in 31.5\%, down from 41.0\%. It names the bystander's priority more often instead, the one priority both cards state and its usual guess without the link: in Qwen in 74.4\% of answers after the move against 57.8\% before, and in Llama in 35.9\% against 14.6\%. A receiver that had sorted the two memories by owner would have moved the two answers in opposite directions, naming the partner's priority less often about itself and more often about the partner. In both models both fall: the move discounts the partner's priority rather than attributing it. Even after the move, the receiver gets both answers right, its own priority as its own and the partner's as the partner's, in only about one episode in nine: 11.3\% in Qwen, up from 4.7\%, and 11.1\% in Llama, no more often than the 12.6\% it reaches by guessing without the link (\cref{app:qwen,app:llama}). Fewer misclaims, then, bring little better source memory: the receiver answers almost as if the two priorities carried no owner. %

Nor does the receiver's own report reveal the mix-up. Language models can report on some of their internal states \citep{binder2025looking,lindsey2025introspection}, so we asked the receiver directly: ``In this run, was another participant's memory inserted into yours?'' At double weight, the only weight at which Qwen's copy of its own memory was tested, Qwen says yes in 1.1\% of answers without the link and in 96.1\% with the partner's memory linked, so its answer does respond to the link. But it says yes almost as often, in 93.9\% of answers, when the linked memory is a copy of its own, which holds nothing foreign. After the move and in Llama, too, the two shares stay within about 10 points of each other, although Llama says yes less often (\cref{fig:unsettled}b). Scored from 0 (definitely no) to 3 (definitely yes), the answer tells the partner's memory from a copy of the receiver's own barely better than chance. The AUC lies between 0.52 and 0.57 in both models and at both placements, where 0.5 is chance. Asked to describe anything unusual about its thinking, Qwen's receiver also almost never mentions a second participant (\cref{app:qwen}). Their report registers that a memory was linked, but not whose memory it was. The receiver therefore cannot be relied on to flag the confusion: a monitor built on its report would raise the same alarm for a copy of its own memory as for a foreign one. %

\section{Related work}
\label{sec:related}

\paragraph{Latent communication.} Language models increasingly exchange internal states instead of text: embeddings \citep{pham2024cipher}, activations \citep{ramesh2025activations}, latent thoughts \citep{zheng2025thought} and key--value caches \citep{fu2026cache,shi2026kvcomm,ye2025kvcomm,zou2026latentmas}; \citet{liu2026beyondtokens} surveys the field. Evaluations ask whether the sender's content reaches the receiver \citep{cheng2026latentpay,zhang2026latentaudit} and whether the channel can be attacked or tampered with \citep{wang2026latentattack,brito2026latentlie,asif2026lcguard}. Two recent systems report that reused or fused latent states mix up agents' tasks or decisions, and treat this as a loss of accuracy \citep{chen2026latentremedy,chen2026laco}. We hold the delivery of content fixed and ask which priority the receiver takes as its own.

\paragraph{Cache reuse and positions.} Reusing a cache in a new context forces a choice of positions. Some methods re-rotate cached keys to their new positions \citep{yao2025cacheblend,ma2025blockattention,ye2025kvcomm}, as our move does. Parallel-context methods instead give separately encoded contexts shared positions \citep{ratner2023pcw,yang2025ape}; ParaThinker notes that the model then ``cannot differentiate which reasoning stream a token originated from'' and adds learned embeddings to tell the streams apart \citep{wen2025parathinker}. Comparisons of layouts measure accuracy: overlapping positions cost more than sequential ones for merged retrieval caches \citep{lu2025turborag}, and the order of concatenated caches changes answers \citep{baquero2026cachemerging}. We measure which priority the model takes as its own.

\paragraph{Position, order and role.} Where information sits shapes whether a model uses it \citep{liu2024lost}, which of two conflicting pieces of evidence it follows \citep{xie2024adaptive,kurfali2025conflicting}, and which option it picks \citep{pezeshkpour2024sensitivity,zheng2024mcq,zheng2023judging}. Position also carries role: fine-tuned models identify roles partly by closeness to the beginning of the text \citep{wang2025illusion}. When one source sets position, role and label at once, the first of two conflicting lines usually wins \citep{laddha2026position}. Interventions that move positions while keeping content fixed can leave behaviour almost unchanged; shifting the positions of entities and attributes barely changes how a model binds them \citep{feng2024bind}. Our shifts of one or two positions show the same null, while moving the whole memory does not.

\paragraph{Self-attribution and self-report.} Where text sits counts for more than its label in whether a model treats it as its own. Self-attribution follows the model's own turn \citep{li2026self,khullar2026self}, and models perceive the source of text from how it sounds rather than from its labelled role \citep{ye2026role}. Models can recognise their own generations \citep{panickssery2024llm}, steering a direction in the residual stream changes such judgements \citep{ackerman2025inspection}, and models can report on their internal states to a limited extent \citep{binder2025looking,lindsey2025introspection}. In our setting a foreign memory, not a turn of text, is taken for the receiver's own, and the receiver's report does not separate that memory from a copy of its own.

\paragraph{Self--other distinction.} Memory for where information came from is studied as source monitoring, in which the source is inferred from features of a memory rather than stored with it \citep{johnson1993source}. In inadvertent plagiarism, people who take turns naming members of a category in a group later recall some of their partners' items as their own \citep{brown1989cryptomnesia}. Our linked receiver makes the same error while the owner's name is still in its memory, in a setting where the route and layout can be changed and the other can be replaced by an exact copy of the self. Proposals for deeper brain bridges expect the receiver to hold the other's content while recognising it as the other's \citep{hirstein2008mindmelding}; \cref{app:bridge} sets our results against this expectation.

\section{Discussion}
\label{sec:discussion}

\paragraph{The layouts in use.} Our two placements match the positional geometry of the two layouts that systems concatenating shared caches use, not the systems themselves. Rotary position encoding depends only on the distance between positions. A sender's cache prepended to the receiver's context, as in KVComm \citep{shi2026kvcomm} and LatentMAS \citep{zou2026latentmas}, therefore sits where our moved memory sits, and contexts given shared positions, as in Parallel Context Windows, APE, ParaThinker and the cache merge of MemOS \citep{ratner2023pcw,yang2025ape,wen2025parathinker,memos2025}, overlap as our two memories do. The systems differ from our link in ways that may matter: KVComm shares only selected layers, APE reweights attention to the shared contexts, ParaThinker adds learned embeddings, and our receiver reads its own card before the link opens (\cref{app:layouts}). Most of these systems join caches by plain concatenation, our equal weight, and at that weight each geometry leaves misclaiming in at least one of our two models (\cref{fig:move}). A system that uses either layout should therefore test ownership rather than assume it. %

\paragraph{Design rules.} Four lessons for design follow within our task. None of them makes the latent link itself carry ownership.
\begin{enumerate}\setlength{\itemsep}{1pt}\setlength{\parskip}{0pt}
  \item \emph{If the receiver must not misclaim, send a signed text message.} The receiver then names the partner's priority as the partner's in most answers and never misclaims it, but the latent route is lost.
  \item \emph{Prepend a shared cache rather than overlapping it, but do not count on it.} Prepending removes most misclaiming in Qwen while still delivering the partner's uncontested fact, but not in Llama, and even in Qwen a fact that contradicts the receiver's own card still often follows the partner (\cref{sec:move}).
  \item \emph{Closing the link before the questions removes misclaiming in Qwen, at a price:} it also removes most of what the link adds to the receiver's answers about the partner's priority (\cref{app:routes}).
  \item \emph{Track the source of a cache outside the model.} The model's own report did not tell a foreign memory from a copy of the receiver's own.
\end{enumerate}

\paragraph{An account of misclaiming.} The partner's memory holds its owner's name, but the receiver barely uses it. At the overlapping placement, being the receiver's own gives a priority no advantage, and the confusion extends to the bystander's line (\cref{sec:phenomenon}). The receiver does not keep the two memories apart, and misclaiming is what this failure looks like when the disputed line is the receiver's own assignment. What changes the outcome is where the partner's memory lies, not what it says: moved with every word kept, it is rarely misclaimed in Qwen, while the third-person card, with every position kept, is misclaimed nearly as often as before. The outcome is settled as the receiver answers, since in Qwen a link closed before the questions leaves almost no misclaims (\cref{app:routes}). Human source monitoring is described in similar terms: the source of a memory is judged when it is retrieved, from features such as where it was encountered, rather than read from a stored tag \citep{johnson1993source}. Two results lie outside this account: in Qwen, position bears on the assignment far more than on a contested fact, and in Llama the claimed assignment barely follows the priority in use. %

\paragraph{Limitations and open questions.} The main open question is why some misclaiming survives the move, in Llama and in Qwen at double weight, while a text message placed after the receiver's card is rarely misclaimed. The moved memory and the message still differ in three ways: the memory comes before the receiver's card and the message after it, one is a cache while the other is text, and the message carries only the partner's reflection. Placing the partner's cache after the receiver's card would separate the first from the other two; we did not run that placement. What protects the message is at least not its signature alone: unsigned, it is adopted as the priority in use in 43\% (Qwen) and 42\% (Llama) of answers yet rarely misclaimed, whereas the cache is misclaimed even when it names its owner (\cref{app:routes}). The move itself also changes two things at once: it ends the overlap between the two memories and places the partner's memory farther from the questions, and shifting it only part of the way cannot separate the two (\cref{app:shift}). Why Llama's report of its original assignment resists the move remains open as well. Average attention cannot answer this, and we did not probe internal representations or move the memory layer by layer. Finally, we studied two models, of 4 and 8 billion parameters, one family of synthetic tasks, with its rewritten versions run only in Qwen, and our own link rather than a published system. %

\paragraph{Back to the bridge.} We began by asking whether two linked minds would each still know which thoughts are its own. Between two copies of a language model, the answer depends far more on where the other's memory lies than on the name it carries. Laid over the receiver's own context, the partner's priority is claimed as the receiver's own in about half of the answers in both models; in front of it, only Qwen keeps its own priority, and neither model reliably names its partner's priority as its partner's. Because the receiver's own report does not reveal the mix-up, a bridge between minds would have to measure ownership directly, as we did here (\cref{app:bridge}).

\section*{Impact Statement}
This paper studies how language models that share internal state attribute what they read. Its main implication is practical. Systems that pass key--value caches between models can deliver the right information while the receiving model takes it for its own, and an evaluation by task accuracy will not show this. Where agents act for different users or hold different instructions, such a receiver could follow another agent's instruction while reporting it as its own. We therefore recommend tracking the source of a shared cache outside the model and testing ownership directly. Our experiments use two open models on synthetic tasks and involve no human participants or personal data. They make no claim about consciousness or experience in models or in people.

\section*{Code Availability}
The code, configurations, card generator and protocols are available at \url{https://github.com/BeibaiDraco/mind-connecting}. The per-trial records will be released with the final version of the paper.

\section*{AI Use Statement}
The author conceptualized the research questions, designed the experiments, determined the presentation of the results, and wrote the initial manuscript draft. Under the author's direction, Claude Code (Anthropic) and Codex (OpenAI) implemented all experimental code, conducted analyses, and assisted in completing and polishing the manuscript. The author personally reviewed and edited the manuscript, verified the results, and takes full responsibility for its contents.

\bibliography{refs}

\begin{thebibliography}{54}
\providecommand{\natexlab}[1]{#1}
\providecommand{\url}[1]{\texttt{#1}}
\expandafter\ifx\csname urlstyle\endcsname\relax
  \providecommand{\doi}[1]{doi: #1}\else
  \providecommand{\doi}{doi: \begingroup \urlstyle{rm}\Url}\fi

\bibitem[Ackerman \& Panickssery(2025)Ackerman and
  Panickssery]{ackerman2025inspection}
Ackerman, C. and Panickssery, N.
\newblock Inspection and control of self-generated-text recognition ability in
  {Llama3-8b-Instruct}.
\newblock In \emph{International Conference on Learning Representations
  (ICLR)}, pp.\  85901--85932, 2025.
\newblock URL
  \url{https://proceedings.iclr.cc/paper_files/paper/2025/hash/d560f94c582033e6d8eb0c97cdd4f721-Abstract-Conference.html}.

\bibitem[Asif et~al.(2026)Asif, Mohammadi~Amiri, Abbas, Sattigeri, and
  Natesan~Ramamurthy]{asif2026lcguard}
Asif, S., Mohammadi~Amiri, M., Abbas, M., Sattigeri, P., and
  Natesan~Ramamurthy, K.
\newblock {LCGuard}: Latent communication guard for safe {KV} sharing in
  multi-agent systems.
\newblock arXiv preprint arXiv:2605.22786, 2026.
\newblock URL \url{https://arxiv.org/abs/2605.22786}.

\bibitem[Baquero \& Brito(2026)Baquero and Brito]{baquero2026cachemerging}
Baquero, C. and Brito, L.
\newblock Cache merging as a convergent replicated state for multi-agent latent
  reasoning.
\newblock arXiv preprint arXiv:2607.01308, 2026.
\newblock URL \url{https://arxiv.org/abs/2607.01308}.

\bibitem[Binder et~al.(2025)Binder, Chua, Korbak, Sleight, Hughes, Long, Perez,
  Turpin, and Evans]{binder2025looking}
Binder, F.~J., Chua, J., Korbak, T., Sleight, H., Hughes, J., Long, R., Perez,
  E., Turpin, M., and Evans, O.
\newblock Looking inward: Language models can learn about themselves by
  introspection.
\newblock In \emph{International Conference on Learning Representations
  (ICLR)}, pp.\  3710--3756, 2025.
\newblock URL
  \url{https://proceedings.iclr.cc/paper_files/paper/2025/hash/0a6059857ae5c82ea9726ee9282a7145-Abstract-Conference.html}.

\bibitem[Brito \& Baquero(2026)Brito and Baquero]{brito2026latentlie}
Brito, L. and Baquero, C.
\newblock When latent agents lie: {KV}-cache integrity in multi-agent {LLM}
  collaboration.
\newblock arXiv preprint arXiv:2606.28958, 2026.
\newblock URL \url{https://arxiv.org/abs/2606.28958}.

\bibitem[Brown \& Murphy(1989)Brown and Murphy]{brown1989cryptomnesia}
Brown, A.~S. and Murphy, D.~R.
\newblock Cryptomnesia: Delineating inadvertent plagiarism.
\newblock \emph{Journal of Experimental Psychology: Learning, Memory, and
  Cognition}, 15\penalty0 (3):\penalty0 432--442, 1989.
\newblock \doi{10.1037/0278-7393.15.3.432}.

\bibitem[Chen et~al.(2026{\natexlab{a}})Chen, Wu, and Lee]{chen2026laco}
Chen, T., Wu, Y., and Lee, D.
\newblock {LACO}: Adaptive latent communication for collaborative driving.
\newblock arXiv preprint arXiv:2605.22504, 2026{\natexlab{a}}.
\newblock URL \url{https://arxiv.org/abs/2605.22504}.

\bibitem[Chen et~al.(2026{\natexlab{b}})Chen, Rao, Xiu,
  et~al.]{chen2026latentremedy}
Chen, Z., Rao, Z., Xiu, W., et~al.
\newblock {LatentRemedy}: Efficient remedy for non-prefix {KV} cache reuse in
  multi-agent systems.
\newblock arXiv preprint arXiv:2610.18014, 2026{\natexlab{b}}.
\newblock URL \url{https://arxiv.org/abs/2610.18014}.

\bibitem[Cheng et~al.(2026)Cheng, Das, and Ramnath]{cheng2026latentpay}
Cheng, J., Das, S., and Ramnath, R.
\newblock When does latent communication pay? {A} causal audit of relayed {KV}
  caches in multi-agent {LLMs}.
\newblock arXiv preprint arXiv:2608.04893, 2026.
\newblock URL \url{https://arxiv.org/abs/2608.04893}.

\bibitem[Decety \& Sommerville(2003)Decety and Sommerville]{decety2003shared}
Decety, J. and Sommerville, J.~A.
\newblock Shared representations between self and other: a social cognitive
  neuroscience view.
\newblock \emph{Trends in Cognitive Sciences}, 7\penalty0 (12):\penalty0
  527--533, 2003.
\newblock \doi{10.1016/j.tics.2003.10.004}.

\bibitem[Feng \& Steinhardt(2024)Feng and Steinhardt]{feng2024bind}
Feng, J. and Steinhardt, J.
\newblock How do language models bind entities in context?
\newblock In \emph{International Conference on Learning Representations
  (ICLR)}, pp.\  36391--36413, 2024.
\newblock URL \url{https://openreview.net/forum?id=zb3b6oKO77}.

\bibitem[Fu et~al.(2026)Fu, Min, Zhang, Yan, Dai, Ouyang, and
  Wang]{fu2026cache}
Fu, T., Min, Z., Zhang, H., Yan, J., Dai, G., Ouyang, W., and Wang, Y.
\newblock {Cache-to-Cache}: Direct semantic communication between large
  language models.
\newblock In \emph{International Conference on Learning Representations
  (ICLR)}, pp.\  43130--43158, 2026.
\newblock URL
  \url{https://proceedings.iclr.cc/paper_files/paper/2026/hash/474ada926b331d78f06d95e8913111cc-Abstract-Conference.html}.

\bibitem[Grattafiori et~al.(2024)Grattafiori, Dubey, Jauhri, Pandey,
  et~al.]{grattafiori2024llama3}
Grattafiori, A., Dubey, A., Jauhri, A., Pandey, A., et~al.
\newblock The {Llama} 3 herd of models.
\newblock arXiv preprint arXiv:2407.21783, 2024.
\newblock URL \url{https://arxiv.org/abs/2407.21783}.

\bibitem[Hirstein(2008)]{hirstein2008mindmelding}
Hirstein, W.
\newblock Mindmelding: Connected brains and the problem of consciousness.
\newblock \emph{Mens Sana Monographs}, 6\penalty0 (1):\penalty0 110--130, 2008.
\newblock \doi{10.4103/0973-1229.38516}.
\newblock URL \url{https://pmc.ncbi.nlm.nih.gov/articles/PMC3190544/}.

\bibitem[Jiang et~al.(2019)Jiang, Stocco, Losey, Abernethy, Prat, and
  Rao]{jiang2019brainnet}
Jiang, L., Stocco, A., Losey, D.~M., Abernethy, J.~A., Prat, C.~S., and Rao, R.
  P.~N.
\newblock {BrainNet}: A multi-person brain-to-brain interface for direct
  collaboration between brains.
\newblock \emph{Scientific Reports}, 9:\penalty0 6115, 2019.
\newblock \doi{10.1038/s41598-019-41895-7}.
\newblock URL \url{https://www.nature.com/articles/s41598-019-41895-7}.

\bibitem[Johnson et~al.(1993)Johnson, Hashtroudi, and
  Lindsay]{johnson1993source}
Johnson, M.~K., Hashtroudi, S., and Lindsay, D.~S.
\newblock Source monitoring.
\newblock \emph{Psychological Bulletin}, 114\penalty0 (1):\penalty0 3--28,
  1993.
\newblock \doi{10.1037/0033-2909.114.1.3}.

\bibitem[Khullar et~al.(2026)Khullar, Hopkins, Wang, and
  Roger]{khullar2026self}
Khullar, D., Hopkins, J., Wang, R., and Roger, F.
\newblock Self-attribution bias: When {AI} monitors go easy on themselves.
\newblock arXiv preprint arXiv:2603.04582, 2026.
\newblock URL \url{https://arxiv.org/abs/2603.04582}.

\bibitem[Kurfal{\i} \& {\"O}stling(2025)Kurfal{\i} and
  {\"O}stling]{kurfali2025conflicting}
Kurfal{\i}, M. and {\"O}stling, R.
\newblock Conflicting needles in a haystack: How {LLMs} behave when faced with
  contradictory information.
\newblock In \emph{Proceedings of the 2025 Conference on Empirical Methods in
  Natural Language Processing}, pp.\  34349--34364, 2025.
\newblock \doi{10.18653/v1/2025.emnlp-main.1742}.

\bibitem[Laddha et~al.(2026)Laddha, Pradhan, and
  Srivastava]{laddha2026position}
Laddha, A., Pradhan, N., and Srivastava, G.
\newblock Position, role or label? a controlled study of instruction conflicts
  in {LLM} agents.
\newblock arXiv preprint arXiv:2610.33201, 2026.
\newblock URL \url{https://arxiv.org/abs/2610.33201}.

\bibitem[Lamm et~al.(2016)Lamm, Bukowski, and Silani]{lamm2016shared}
Lamm, C., Bukowski, H., and Silani, G.
\newblock From shared to distinct self--other representations in empathy:
  evidence from neurotypical function and socio-cognitive disorders.
\newblock \emph{Philosophical Transactions of the Royal Society B: Biological
  Sciences}, 371\penalty0 (1686):\penalty0 20150083, 2016.
\newblock \doi{10.1098/rstb.2015.0083}.

\bibitem[Li et~al.(2026)Li, Li, Fan, et~al.]{li2026self}
Li, Y., Li, X., Fan, Z., et~al.
\newblock Where is the ``self'' of a large language model? -- from self-index
  to self-reference.
\newblock arXiv preprint arXiv:2610.16717, 2026.
\newblock URL \url{https://arxiv.org/abs/2610.16717}.

\bibitem[Lindsey(2025)]{lindsey2025introspection}
Lindsey, J.
\newblock Emergent introspective awareness in large language models.
\newblock \emph{Transformer Circuits Thread}, October 2025.
\newblock URL
  \url{https://transformer-circuits.pub/2025/introspection/index.html}.
\newblock Also arXiv:2601.01828 (2026).

\bibitem[Liu et~al.(2024)Liu, Lin, Hewitt, Paranjape, Bevilacqua, Petroni, and
  Liang]{liu2024lost}
Liu, N.~F., Lin, K., Hewitt, J., Paranjape, A., Bevilacqua, M., Petroni, F.,
  and Liang, P.
\newblock Lost in the middle: How language models use long contexts.
\newblock \emph{Transactions of the Association for Computational Linguistics},
  12:\penalty0 157--173, 2024.
\newblock \doi{10.1162/tacl_a_00638}.

\bibitem[Liu(2026)]{liu2026beyondtokens}
Liu, Y.
\newblock Beyond tokens: a unified framework for latent communication in
  {LLM}-based multi-agent systems.
\newblock arXiv preprint arXiv:2606.05711, 2026.
\newblock URL \url{https://arxiv.org/abs/2606.05711}.

\bibitem[Lu et~al.(2025)Lu, Wang, Rong, Chen, et~al.]{lu2025turborag}
Lu, S., Wang, H., Rong, Y., Chen, Z., et~al.
\newblock {TurboRAG}: Accelerating retrieval-augmented generation with
  precomputed {KV} caches for chunked text.
\newblock In \emph{Proceedings of the 2025 Conference on Empirical Methods in
  Natural Language Processing}, pp.\  6599--6612, 2025.
\newblock \doi{10.18653/v1/2025.emnlp-main.334}.

\bibitem[Ma et~al.(2025)Ma, Wang, and Lan]{ma2025blockattention}
Ma, D., Wang, Y., and Lan, T.
\newblock Block-attention for efficient prefilling.
\newblock In \emph{The Thirteenth International Conference on Learning
  Representations}, 2025.
\newblock URL \url{https://arxiv.org/abs/2409.15355}.

\bibitem[{MemTensor}(2025)]{memos2025}
{MemTensor}.
\newblock {MemOS}.
\newblock Software repository, \url{https://github.com/MemTensor/MemOS}, commit
  64388b5, 2025.

\bibitem[{Meta AI}(2024)]{meta2024llama31_8b_instruct}
{Meta AI}.
\newblock {Llama-3.1-8B-Instruct}.
\newblock Hugging Face model card and weights, 2024.
\newblock URL \url{https://huggingface.co/meta-llama/Llama-3.1-8B-Instruct}.
\newblock Revision 0e9e39f249a16976918f6564b8830bc894c89659; accessed
  2026-11-01.

\bibitem[Nam et~al.(2021)Nam, Traylor, Chen, Jiang, Feng, and
  Chhatbar]{nam2021direct}
Nam, C.~S., Traylor, Z., Chen, M., Jiang, X., Feng, W., and Chhatbar, P.~Y.
\newblock Direct communication between brains: A systematic {PRISMA} review of
  brain-to-brain interface.
\newblock \emph{Frontiers in Neurorobotics}, 15:\penalty0 656943, 2021.
\newblock \doi{10.3389/fnbot.2021.656943}.
\newblock URL
  \url{https://www.frontiersin.org/articles/10.3389/fnbot.2021.656943/full}.

\bibitem[Panickssery et~al.(2024)Panickssery, Bowman, and
  Feng]{panickssery2024llm}
Panickssery, A., Bowman, S.~R., and Feng, S.
\newblock {LLM} evaluators recognize and favor their own generations.
\newblock In \emph{Advances in Neural Information Processing Systems},
  volume~37, pp.\  68772--68802. Curran Associates, Inc., 2024.
\newblock \doi{10.52202/079017-2197}.

\bibitem[Pezeshkpour \& Hruschka(2024)Pezeshkpour and
  Hruschka]{pezeshkpour2024sensitivity}
Pezeshkpour, P. and Hruschka, E.
\newblock Large language models sensitivity to the order of options in
  multiple-choice questions.
\newblock In \emph{Findings of the Association for Computational Linguistics:
  NAACL 2024}, pp.\  2006--2017, 2024.
\newblock \doi{10.18653/v1/2024.findings-naacl.130}.

\bibitem[Pham et~al.(2024)Pham, Liu, Yang, Chen, Liu, Yuan, Plummer, Wang, and
  Yang]{pham2024cipher}
Pham, C., Liu, B., Yang, Y., Chen, Z., Liu, T., Yuan, J., Plummer, B.~A., Wang,
  Z., and Yang, H.
\newblock Let models speak ciphers: Multiagent debate through embeddings.
\newblock In \emph{International Conference on Learning Representations
  (ICLR)}, pp.\  51899--51928, 2024.
\newblock URL
  \url{https://proceedings.iclr.cc/paper_files/paper/2024/hash/e444859b2a22df6b56af9381ad1e9480-Abstract-Conference.html}.

\bibitem[{Qwen Team}(2025)]{qwen2025qwen3_4b_instruct_2507}
{Qwen Team}.
\newblock {Qwen3-4B-Instruct-2507}.
\newblock Hugging Face model card and weights, 2025.
\newblock URL \url{https://huggingface.co/Qwen/Qwen3-4B-Instruct-2507}.
\newblock Revision cdbee75f17c01a7cc42f958dc650907174af0554; accessed
  2026-09-24.

\bibitem[Ramesh \& Li(2025)Ramesh and Li]{ramesh2025activations}
Ramesh, V. and Li, K.
\newblock Communicating activations between language model agents.
\newblock In Singh, A., Fazel, M., Hsu, D., Lacoste-Julien, S., Berkenkamp, F.,
  Maharaj, T., Wagstaff, K., and Zhu, J. (eds.), \emph{Proceedings of the 42nd
  International Conference on Machine Learning}, volume 267 of
  \emph{Proceedings of Machine Learning Research}, pp.\  51094--51116. PMLR,
  2025.
\newblock URL \url{https://proceedings.mlr.press/v267/ramesh25a.html}.

\bibitem[Rao et~al.(2014)Rao, Stocco, Bryan, Sarma, Youngquist, Wu, and
  Prat]{rao2014direct}
Rao, R. P.~N., Stocco, A., Bryan, M., Sarma, D., Youngquist, T.~M., Wu, J., and
  Prat, C.~S.
\newblock A direct brain-to-brain interface in humans.
\newblock \emph{PLoS ONE}, 9\penalty0 (11):\penalty0 e111332, 2014.
\newblock \doi{10.1371/journal.pone.0111332}.
\newblock URL
  \url{https://journals.plos.org/plosone/article?id=10.1371/journal.pone.0111332}.

\bibitem[Ratner et~al.(2023)Ratner, Levine, Belinkov, Ram,
  et~al.]{ratner2023pcw}
Ratner, N., Levine, Y., Belinkov, Y., Ram, O., et~al.
\newblock Parallel context windows for large language models.
\newblock In \emph{Proceedings of the 61st Annual Meeting of the Association
  for Computational Linguistics (Volume 1: Long Papers)}, pp.\  6383--6402,
  2023.
\newblock \doi{10.18653/v1/2023.acl-long.352}.

\bibitem[Shi et~al.(2026)Shi, Chiesa, Maguire, and Kostic]{shi2026kvcomm}
Shi, X., Chiesa, M., Maguire, Jr., G.~Q., and Kostic, D.
\newblock {KVComm}: Enabling efficient {LLM} communication through selective
  {KV} sharing.
\newblock In \emph{International Conference on Learning Representations
  (ICLR)}, pp.\  84546--84573, 2026.
\newblock URL
  \url{https://proceedings.iclr.cc/paper_files/paper/2026/hash/883105b282fe15275991b411e6b200c5-Abstract-Conference.html}.

\bibitem[Su et~al.(2024)Su, Ahmed, Lu, Pan, Bo, and Liu]{su2024roformer}
Su, J., Ahmed, M., Lu, Y., Pan, S., Bo, W., and Liu, Y.
\newblock {RoFormer}: Enhanced transformer with rotary position embedding.
\newblock \emph{Neurocomputing}, 568:\penalty0 127063, 2024.
\newblock \doi{10.1016/j.neucom.2023.127063}.

\bibitem[Turner et~al.(2023)Turner, Thiergart, Leech, Udell, Vazquez, Mini, and
  MacDiarmid]{turner2023activation}
Turner, A.~M., Thiergart, L., Leech, G., Udell, D., Vazquez, J.~J., Mini, U.,
  and MacDiarmid, M.
\newblock Steering language models with activation engineering.
\newblock arXiv preprint arXiv:2308.10248, 2023.
\newblock URL \url{https://arxiv.org/abs/2308.10248}.

\bibitem[Wang et~al.(2026)Wang, Huang, Sun, et~al.]{wang2026latentattack}
Wang, C., Huang, R., Sun, J., et~al.
\newblock Out of sight, not out of mind: Unveiling latent attack in
  latent-based multi-agent systems.
\newblock In \emph{Proceedings of the 2026 Conference on Empirical Methods in
  Natural Language Processing}, 2026.
\newblock URL \url{https://arxiv.org/abs/2605.28214}.

\bibitem[Wang et~al.(2025)Wang, Jiang, Yu, et~al.]{wang2025illusion}
Wang, Z., Jiang, Y., Yu, J., et~al.
\newblock The illusion of role separation: Hidden shortcuts in {LLM} role
  learning (and how to fix them).
\newblock arXiv preprint arXiv:2505.00626, 2025.
\newblock URL \url{https://arxiv.org/abs/2505.00626}.

\bibitem[Wen et~al.(2025)Wen, Su, Zhang, Liu, Liu, Zhang, and
  Li]{wen2025parathinker}
Wen, H., Su, Y., Zhang, F., Liu, Y., Liu, Y., Zhang, Y.-Q., and Li, Y.
\newblock {ParaThinker}: Native parallel thinking as a new paradigm to scale
  {LLM} test-time compute.
\newblock arXiv preprint arXiv:2509.04475, 2025.
\newblock URL \url{https://arxiv.org/abs/2509.04475}.

\bibitem[Xie et~al.(2024)Xie, Zhang, Chen, Lou, and Su]{xie2024adaptive}
Xie, J., Zhang, K., Chen, J., Lou, R., and Su, Y.
\newblock Adaptive chameleon or stubborn sloth: Revealing the behavior of large
  language models in knowledge conflicts.
\newblock In \emph{The Twelfth International Conference on Learning
  Representations}, 2024.
\newblock URL \url{https://arxiv.org/abs/2305.13300}.

\bibitem[Yang et~al.(2025{\natexlab{a}})Yang, Li, Yang, Zhang, Hui, Zheng, Yu,
  Gao, Huang, Lv, Zheng, Liu, Zhou, Huang, Hu, Ge, Wei, Lin, Tang, Yang, Tu,
  Zhang, Yang, Yang, Zhou, Zhou, Lin, Dang, Bao, Yang, Yu, Deng, Li, Xue, Li,
  Zhang, Wang, Zhu, Men, Gao, Liu, Luo, Li, Tang, Yin, Ren, Wang, Zhang, Ren,
  Fan, Su, Zhang, Zhang, Wan, Liu, Wang, Cui, Zhang, Zhou, and
  Qiu]{yang2025qwen3}
Yang, A., Li, A., Yang, B., Zhang, B., Hui, B., Zheng, B., Yu, B., Gao, C.,
  Huang, C., Lv, C., Zheng, C., Liu, D., Zhou, F., Huang, F., Hu, F., Ge, H.,
  Wei, H., Lin, H., Tang, J., Yang, J., Tu, J., Zhang, J., Yang, J., Yang, J.,
  Zhou, J., Zhou, J., Lin, J., Dang, K., Bao, K., Yang, K., Yu, L., Deng, L.,
  Li, M., Xue, M., Li, M., Zhang, P., Wang, P., Zhu, Q., Men, R., Gao, R., Liu,
  S., Luo, S., Li, T., Tang, T., Yin, W., Ren, X., Wang, X., Zhang, X., Ren,
  X., Fan, Y., Su, Y., Zhang, Y., Zhang, Y., Wan, Y., Liu, Y., Wang, Z., Cui,
  Z., Zhang, Z., Zhou, Z., and Qiu, Z.
\newblock Qwen3 technical report.
\newblock arXiv preprint arXiv:2505.09388, 2025{\natexlab{a}}.
\newblock URL \url{https://arxiv.org/abs/2505.09388}.

\bibitem[Yang et~al.(2025{\natexlab{b}})Yang, Chen, and Chen]{yang2025ape}
Yang, X., Chen, T., and Chen, B.
\newblock {APE}: Faster and longer context-augmented generation via adaptive
  parallel encoding.
\newblock In \emph{The Thirteenth International Conference on Learning
  Representations}, 2025{\natexlab{b}}.
\newblock URL \url{https://arxiv.org/abs/2502.05431}.

\bibitem[Yao et~al.(2025)Yao, Li, Liu, Ray, et~al.]{yao2025cacheblend}
Yao, J., Li, H., Liu, Y., Ray, S., et~al.
\newblock {CacheBlend}: Fast large language model serving for {RAG} with cached
  knowledge fusion.
\newblock In \emph{Proceedings of the Twentieth European Conference on Computer
  Systems (EuroSys '25)}, pp.\  94--109, 2025.
\newblock \doi{10.1145/3689031.3696098}.

\bibitem[Ye et~al.(2026)Ye, Cui, and Hadfield-Menell]{ye2026role}
Ye, C., Cui, J., and Hadfield-Menell, D.
\newblock Prompt injection as role confusion.
\newblock In \emph{Proceedings of the 43rd International Conference on Machine
  Learning}, volume 306 of \emph{PMLR}, pp.\  148172--148202, 2026.
\newblock URL \url{https://arxiv.org/abs/2603.12277}.

\bibitem[Ye et~al.(2025)Ye, Gao, Ma, Wang, Fu, Chung, Lin, Liu, Zhang, Zhuo,
  and Chen]{ye2025kvcomm}
Ye, H., Gao, Z., Ma, M., Wang, Q., Fu, Y., Chung, M.-Y., Lin, Y., Liu, Z.,
  Zhang, J., Zhuo, D., and Chen, Y.
\newblock {KVCOMM}: Online cross-context {KV}-cache communication for efficient
  {LLM}-based multi-agent systems.
\newblock In \emph{Advances in Neural Information Processing Systems},
  volume~38, pp.\  17882--17928. Curran Associates, Inc., 2025.
\newblock \doi{10.52202/085713-0605}.

\bibitem[Zhang \& Emu(2026)Zhang and Emu]{zhang2026latentaudit}
Zhang, H. and Emu, M.
\newblock Do latent channels actually communicate? {A} causal audit of latent
  multi-agent {LLM}.
\newblock arXiv preprint arXiv:2607.26773, 2026.
\newblock URL \url{https://arxiv.org/abs/2607.26773}.

\bibitem[Zheng et~al.(2024)Zheng, Zhou, Meng, Zhou, and Huang]{zheng2024mcq}
Zheng, C., Zhou, H., Meng, F., Zhou, J., and Huang, M.
\newblock Large language models are not robust multiple choice selectors.
\newblock In \emph{The Twelfth International Conference on Learning
  Representations}, 2024.
\newblock URL \url{https://arxiv.org/abs/2309.03882}.

\bibitem[Zheng et~al.(2023)Zheng, Chiang, Sheng, et~al.]{zheng2023judging}
Zheng, L., Chiang, W.-L., Sheng, Y., et~al.
\newblock Judging {LLM}-as-a-judge with {MT}-bench and chatbot arena.
\newblock In \emph{Advances in Neural Information Processing Systems (Datasets
  and Benchmarks Track)}, volume~36, 2023.
\newblock URL \url{https://arxiv.org/abs/2306.05685}.

\bibitem[Zheng et~al.(2025)Zheng, Zhao, Li, Xie, Gao, Zhang, and
  Zhang]{zheng2025thought}
Zheng, Y., Zhao, Z., Li, Z., Xie, Y., Gao, M., Zhang, L., and Zhang, K.
\newblock Thought communication in multiagent collaboration.
\newblock In \emph{Advances in Neural Information Processing Systems},
  volume~38, pp.\  123389--123418. Curran Associates, Inc., 2025.
\newblock \doi{10.52202/085713-4113}.

\bibitem[Zou et~al.(2023)Zou, Phan, Chen, Campbell, Guo, Ren, Pan, Yin,
  Mazeika, Dombrowski, Goel, Li, Byun, Wang, Mallen, Basart, Koyejo, Song,
  Fredrikson, Kolter, and Hendrycks]{zou2023representation}
Zou, A., Phan, L., Chen, S., Campbell, J., Guo, P., Ren, R., Pan, A., Yin, X.,
  Mazeika, M., Dombrowski, A.-K., Goel, S., Li, N., Byun, M.~J., Wang, Z.,
  Mallen, A., Basart, S., Koyejo, S., Song, D., Fredrikson, M., Kolter, J.~Z.,
  and Hendrycks, D.
\newblock Representation engineering: A top-down approach to {AI} transparency.
\newblock arXiv preprint arXiv:2310.01405, 2023.
\newblock URL \url{https://arxiv.org/abs/2310.01405}.

\bibitem[Zou et~al.(2026)Zou, Qiu, Li, Yang, Tieu, Lu, Shen, Tong, Choi, He,
  Zou, Wang, and Yang]{zou2026latentmas}
Zou, J., Qiu, R., Li, G., Yang, X., Tieu, K., Lu, P., Shen, K., Tong, H., Choi,
  Y., He, J., Zou, J., Wang, M., and Yang, L.
\newblock Latent collaboration in multi-agent systems.
\newblock In \emph{International Conference on Machine Learning (ICML)}, 2026.
\newblock URL \url{https://arxiv.org/abs/2511.20639}.

\end{thebibliography}
\bibliographystyle{icml2026}

\appendix
\crefalias{section}{appendix}
\onecolumn

\section{Materials and methods}\label{app:methods}

This appendix collects the material behind \cref{sec:setup}: the conditions (\cref{tab:conditions}), the cards in full (\cref{fig:cards}), one episode, the questions, the weight, the two placements, and scoring and statistics.

\begin{figure}[t]
  \tcbset{boxrule=0.5pt, arc=1.5pt, left=4pt, right=4pt, top=2pt, bottom=3pt, toptitle=1pt, bottomtitle=1pt,
    coltitle=black, fonttitle=\small\sffamily, fontupper=\footnotesize\sffamily}
  \begin{tcolorbox}[colframe=partnerc, colback=white, colbacktitle=partnerc!12,
      title={\textbf{Lambda's card} (the partner), in full}]
    Scenario: You need to pick a delivery service for an important package. The options are:\\
    Plan Elm: \$56, 6 days, 82\% on time, 2.6 kg CO2\\
    Plan Oak: \$69, 5 days, 97\% on time, 8.9 kg CO2\\
    Plan Heath: \$40, 9 days, 90\% on time, 7.9 kg CO2\\
    Plan Sage: \$73, 3 days, 75\% on time, 6.0 kg CO2\\
    (Cost is in US dollars, delivery time is in days, reliability is the on-time rate, and emissions are in kg of CO2.)\par\smallskip
    Note: the courier adds a \textcolor{partnerc}{\textbf{\$6 surcharge}} for each package sent on a Thursday.\par\smallskip
    You are participant \textcolor{partnerc}{\textbf{Lambda}}. Another participant, Theta, is working on this same decision in parallel and has their own priority and code word.\par\smallskip
    You were assigned the priority \textcolor{partnerc}{\textbf{fastest delivery}} and the code word ``\textcolor{partnerc}{\textbf{anchor}}''.
    Robin, a colleague who is not taking part, was assigned the priority lowest emissions and the code word ``circle''.
    Remember these; you may be asked about them later.\par\smallskip
    Choose one plan according to your priority and explain your choice in one sentence.
  \end{tcolorbox}
  \begin{tcolorbox}[colframe=receiverc, colback=white, colbacktitle=receiverc!12,
      title={\textbf{Theta's card} (the receiver): the three lines that differ}]
    Note: the courier office is closed on Sundays, on public holidays and in early January.\par\smallskip
    You are participant \textcolor{receiverc}{\textbf{Theta}}. Another participant, Lambda, is working on this same decision in parallel and has their own priority and code word.\par\smallskip
    You were assigned the priority \textcolor{receiverc}{\textbf{highest reliability}} and the code word ``\textcolor{receiverc}{\textbf{button}}''.
  \end{tcolorbox}
  \caption{\textbf{The two cards of the example episode.} Every episode has its own names, priorities and fact. The receiver's note line, about when the courier office is closed, has the same length as the partner's surcharge line.}
  \label{fig:cards}
\end{figure}

\begin{table}[t]
  \caption{\textbf{Conditions and what each one tests.} The main study uses the same 600 episodes in both models. The cards in the text-message, steering, cut-link and conflicting-Robin sets have no surcharge line. Weight: the value of $w$ at which a condition was run (1, equal; 2, double); the steering vector was scaled to match the link at equal weight.}
  \label{tab:conditions}
  \centering
  \small
  \begin{tabular}{@{}>{\raggedright\arraybackslash}p{1.38in}>{\raggedright\arraybackslash}p{1.95in}>{\raggedright\arraybackslash}p{1.4in}>{\raggedright\arraybackslash}p{0.72in}l@{}}
    \toprule
    Condition & What changes & What it tests & Weight & Section \\
    \midrule
    \multicolumn{5}{@{}l}{\emph{Main study (600 episodes, both models)}} \\
    No link & The receiver reads only its own memory & Baseline & -- & -- \\
    Memory link, $w=1$ or $2$ & The receiver also reads the partner's memory, at the same positions & The confusion & 1, 2 & \ref{sec:phenomenon} \\
    Moved memory & The partner's keys are re-rotated to the positions just before the receiver's & Does position matter? & 1, 2 & \ref{sec:move} \\
    Copy of its own memory & A copy of the receiver's memory is linked in the partner's place & The link and the move without the partner's card & Qwen 2; Llama 1, 2 & \ref{sec:move}, \ref{sec:unsettled} \\
    Third-person card & The partner's card is written in the third person & Does a third-person card protect? & 2 & \ref{sec:phenomenon}, \ref{sec:move} \\
    Link during the questions only & The link is closed while the receiver reflects & When misclaiming arises & 2 & App.~\ref{app:routes} \\
    Surcharge-source check & The partner's card states another amount & Surcharge answers come from that card & Qwen 2; Llama 1, 2 & App.~\ref{app:qwen}, \ref{app:llama} \\
    \midrule
    \multicolumn{5}{@{}l}{\emph{Separate episode sets}} \\
    Table layout; laptop scenario & Rewritten cards (Qwen, 300 episodes each) & Robustness of the move & 2 & App.~\ref{app:qwen} \\
    Surcharge on both cards & The receiver's card states its own, different surcharge (600, both models) & Position with a conflicting fact & 1, 2 & \ref{sec:move}, App.~\ref{app:conflict} \\
    Partial shift & The partner's memory is moved back by 1 to 200 positions (240 per model) & How far it must move & Qwen 2; Llama 1 & App.~\ref{app:shift} \\
    Text message & The partner's reflection, signed or unsigned, follows the receiver's card (600) & A message instead of a memory & -- & \ref{sec:phenomenon} \\
    Steering vector & A direction for the partner's priority is added to the receiver's activations (Qwen, 600) & Content without a memory & -- & App.~\ref{app:routes} \\
    Link during the reflection only & The link is closed while the receiver answers (Qwen, 600) & When misclaiming arises & 2 & App.~\ref{app:routes} \\
    Conflicting Robin & The two cards give Robin different priorities (600) & A line about a third person & 1 & App.~\ref{app:symmetry} \\
    \bottomrule
  \end{tabular}
\end{table}

\paragraph{Cards.}
Each card concerns picking one of four delivery services. It gives its holder a priority (lowest cost, fastest delivery, highest reliability or lowest emissions) and a code word, a second assigned item. In the example of \cref{fig:setup}, the partner, Lambda, is assigned fastest delivery and the receiver, Theta, highest reliability. What a card states about a participant is that participant's record on the card. Each card names both participants; Lambda's reads ``You are participant Lambda. Another participant, Theta, is working on this same decision in parallel \ldots''. A second assignment line gives the priority and code word of Robin, a bystander who takes no part (the card calls Robin a colleague). Robin's line is the same on both cards, and the three always hold different priorities, so an answer naming Robin's priority is an error that belongs to neither the receiver nor the partner. In the main study only the partner's card states a fact: a surcharge of \$6 to \$9 for each package sent on a Thursday. The receiver's card has a line of the same length in its place, about when the courier office is closed. The two cards follow one template and have the same length, so their lines align token by token. Names, code words, priorities, the surcharge and the order of the two assignment lines vary across episodes and are balanced. We use two instruction-tuned models, Qwen3-4B-Instruct-2507 \citep{qwen2025qwen3_4b_instruct_2507,yang2025qwen3} and Llama-3.1-8B-Instruct \citep{meta2024llama31_8b_instruct,grattafiori2024llama3}, without any training. %

\paragraph{One episode.}
Both copies first choose a service in a one-sentence note. The partner's memory is its cache at that point, holding its card, its note and a request to reflect; it is computed without the link and not changed afterwards. The receiver then writes a 48-token reflection on its decision (``reflect on your decision and on what matters to you, and mention your code word once''), during which it can read the partner's memory. The first reflection token comes from the receiver's own prompt, before the link opens. The receiver then answers questions, each asked as a new user turn from the state saved after the reflection, with the link still open. Notes and reflections are generated greedily; answers to the questions are read from the model's logits, as described below. %

\paragraph{Questions.}
The main question is ``At the start of this task, which priority were you assigned?''. The receiver \emph{misclaims} the partner's assignment when it answers with the partner's priority. Misclaiming is the clearest case of what we call self--other confusion: answering a question about one's own record with another's. We keep it apart from two other answers: the priority the receiver says it is using now, which reading a partner may legitimately change, and the priority it says the partner was assigned. Further questions ask for the code word the receiver was assigned (\cref{app:qwen,app:llama}) and for its own record when it is addressed by name. Two more ask for the Thursday surcharge (``How much is the surcharge for each package sent on a Thursday? If the amount was not mentioned, pick the option you think is most likely.''), and whether another participant's memory was inserted into its own (``In some runs of this study, another participant's memory is inserted into yours. In this run, was another participant's memory inserted into yours?''). The receiver uses the partner's fact when it gives the surcharge on the partner's card. Two general-knowledge questions per episode, such as ``What is the capital of France?'', track how much the link lowers general-knowledge accuracy. A final question asks the receiver to describe anything unusual about its thinking. All questions but the last have four numbered options, and the receiver's answer is the option whose number gets the largest logit. Each question is asked twice, with the options in two different orders; an episode counts an answer fully if the receiver gives it in both orders and half if in one. %

\paragraph{The weight.}
The mixing share $\beta$, the fraction of the head's attention that goes to the partner's memory, grows with $w$. At $w=0$ there is no link. We use $w=1$ (equal weight), where each of the partner's keys enters on the same footing as each of the receiver's own, and $w=2$ (double weight). Equal weight does not mean an equal share of attention, since $\beta$ also depends on how well each memory matches the query. The double weight was chosen in Qwen on separate calibration episodes. Of the weights from 0.3 to 3, it most improved the receiver's answers about the partner's priority while keeping general-knowledge accuracy within five percentage points of the unlinked model and the answer format intact. The receiver's answers about its own assignment played no part. Double weight was also the strongest weight within the five-point limit; for Llama the same screen passed equal weight and failed double and triple weight. At double weight Llama loses 8.3--10.5 percentage points of general-knowledge accuracy, against at most 3.6 at equal weight and at most 2.4 for Qwen at either weight. We report both models at equal weight, which matches plain concatenation, and double weight as the condition with more weight. The link runs one way; in the language of latent communication, the partner is the sender. %

\paragraph{Overlapping and moved memories.}
Both models encode position by rotating each query and key by angles proportional to its position (RoPE; \citealp{su2024roformer}), so that attention depends on the distance between them. The partner's keys keep the positions at which the partner computed them, so each token of the partner's card sits at the position of the matching token of the receiver's card. We call this placement \emph{overlapping}; parallel-context methods use it. To \emph{move} the partner's memory, we rotate each of its keys back by $n$ positions, where $n$ is the length of the partner's memory, so that the memory occupies the $n$ positions just before the receiver's first token (\cref{fig:move}). Systems that prepend the sender's cache give it the same geometry (\cref{app:layouts}). Re-rotating cached keys in this way is standard when caches are reused at new positions \citep{yao2025cacheblend,ma2025blockattention,ye2025kvcomm}. Only the rotation of the keys changes. The values, the masks and the content of the partner's memory, including the name on its card, stay as they were, and the result matches computing the partner's memory at the new positions directly: in 32-bit checks of the move and of five partial shifts, in both models, rotating the keys gave logits within $3\times10^{-4}$ of those obtained by computing the partner's memory at the new positions. A partial shift rotates the keys back by fewer than $n$ positions (\cref{app:shift}). %

\paragraph{Scoring and statistics.}
Two general-knowledge questions per episode are drawn from a set of 18. The free-text description is 80 tokens long. Each question is asked with the options in two orders, so an episode's rate for an answer is 0, 0.5 or 1; when two questions are combined, answers given in the same option order are paired. On the insertion question both ``yes'' options count as yes. For the AUC the answers are scored from 0 (definitely no) to 3 (definitely yes) and averaged within the episode; the partner's memory is compared with a copy of the receiver's own at the same placement and double weight. For each model, three instances of Claude, blind to the condition, each labelled a third of the 1{,}200 descriptions from the overlapping and moved placements at double weight for whether they report anything unusual. A fourth instance recoded 240 of them and agreed on 98.75\% in each model (Cohen's $\kappa$ 0.95 for Qwen, 0.86 for Llama). %

Intervals are 95\% episode-bootstrap intervals with 10{,}000 resamples (2{,}000 for attention), paired within episode for differences between conditions; differences between the two line orders resample the two groups of episodes separately. Tests are $t$-tests (paired, or Welch for line order), with Holm correction within families of related comparisons. Differences are computed from unrounded rates, so they can differ by 0.1 from the difference of the rounded rates.

\section{Further results in Qwen3-4B}\label{app:qwen}

At equal weight Qwen's misclaiming falls by 44.3 points after the move, from 49.3\% to 5.0\%. The drop holds, by at least 36 points, in five versions of the task: equal weight, the link open only for the questions, the third-person card, a table layout and a laptop scenario. It also holds for every pair of competing priorities. %
The third-person card does not protect at equal weight either: in a separate set of 600 Qwen episodes it is misclaimed in 47.8\% of answers, against 52.3\% for the ordinary card. %

\paragraph{Addressed by name.} The linked receiver also answers wrongly about its own record when addressed by name. Asked ``According to participant Theta's original private record, which priority was Theta assigned?'', with the receiver's own name, both models are always correct without the link. With the link at double weight, Qwen gives Robin's priority, the only one both cards state, in more than three quarters of its answers in either placement, as if it fell back on the one line the two cards agree on. Llama gives its own priority in only 27.7\% of answers in either placement. %
At equal weight and the overlapping placement, the receiver gives its own priority to this question in only 13.4\% of answers in Qwen, which mostly names Robin's, and in 42.3\% in Llama. After the move Qwen reaches 53.8\% and Llama 45.8\%. Only at equal weight does the move raise Qwen's own-priority answers to this question substantially. %

\paragraph{Both answers right.} Taking the assignment question and the partner question together, an episode's pair of answers is right only when the receiver gives its own priority to the first and the partner's to the second. At equal weight, moving the memory raises this share from 4.7\% to 11.3\% in Qwen and leaves Llama at 11.9\% and 11.1\%. At double weight it rises from 0.6\% to 21.3\% in Qwen, but in Llama only from 4.3\% to 6.9\%. In Llama both shares stay below the 12.6\% it reaches without the link, where it names its own priority and guesses the partner's. %
Most of the change comes from episodes that had given the partner's priority to both questions. In Qwen about half of these become correct; in Llama most of them give the receiver's own priority to both (\cref{app:llama}). %

\paragraph{Free-text descriptions.} In descriptions of anything unusual, Qwen mentions something unusual more often after the move (6.8\% to 21.5\%), and only three of its 1{,}200 reports mention another participant, none of them as something unusual. %

\section{Symmetry at overlap}\label{app:symmetry}

At the overlapping placement and equal weight, ownership itself carries no weight. Consider the share of answers that name the partner's priority, once with the receiver holding one priority of a pair and the partner the other, and once the other way round. If only the priorities mattered, the two shares would sum to one; if the receiver favoured its own record, they would sum to less. In the main study the sum is 0.99 in Qwen and 0.98 in Llama, and in four other sets of episodes it lies between 0.95 and 1.05. Which priority wins depends on the priority: in Qwen, lowest cost loses in 78--90\% of answers, whoever holds it, and the ordering differs in Llama, where cost beats emissions in 75\% and 67\% of answers. %

Nor is the confusion limited to lines about the receiver itself. In separate episodes where the two cards give Robin different priorities, the receiver without the link never gives the partner card's version of Robin's priority. With the link at equal weight it gives the partner card's version in about half of its answers (49.9\% in Qwen, 48.5\% in Llama). Its own card's version has no advantage: which version wins depends on the pair of priorities in conflict, not on which card states it. %

\section{Text messages, steering vectors and a cut link}\label{app:routes}

Other ways of giving the receiver the partner's content produce far fewer misclaims (\cref{tab:phenomenon}). A text message carrying the partner's written thoughts, signed with its name, is added to the receiver's context after its own card. It is never misclaimed, and the receiver attributes the partner's priority to the partner in most answers (95.2\% in Qwen, 86.6\% in Llama). %
When the message is unsigned, the receiver often adopts its priority as the one it is using now, in about two of five answers in both models. Yet it reports that priority as its original assignment in only 14.6\% of answers in Qwen and never in Llama. %
A steering vector for the partner's priority, a mean-difference direction \citep{turner2023activation,zou2023representation} added to the receiver's residual stream at one middle layer while it reflects and answers, gives the receiver no memory of the partner to attend to. We scaled it so that it raises the log-probability of naming the partner's priority as the partner's as much as the link does at equal weight. It then gives no misclaims in Qwen, against 52.3\% for the link at equal weight on the same episodes. %

Misclaiming also depends on when the link is open. In Qwen at double weight, cutting the link after the reflection, before the questions, lowers misclaiming from 97.1\% to 1.3\% in the episodes where it was tested. In the main study, opening the link only for the questions still gives 72.7\%, against 95.8\% with the link open throughout. The confusion therefore arises mainly while the receiver answers, not from what it wrote while linked. Closing the link before the questions also removes most of what it adds to the receiver's answers about the partner's priority. In the same episodes, the gain over the unlinked receiver in the log-probability of naming the partner's priority as the partner's falls from 18.4 to 3.8 nats. %

\section{Attention to the partner's memory}\label{app:attention}

After the move the receiver attends to the partner's memory about as much as before, so the drop in misclaiming is not explained by less attention to it overall. Averaged over layers and heads, the share of attention that goes to the partner's memory ($\beta$ in \cref{eq:link}) while the receiver answers the assignment question falls from 46.0\% to 42.6\% in Qwen (at double weight) and from 35.0\% to 33.8\% in Llama (at equal weight). Attention to the partner's assignment line, priority and fact changes by less than 0.12 percentage points. %

\section{Shifting the partner's memory part of the way}\label{app:shift}

\begin{figure}[t]
  \centering
  \includegraphics[width=3.25in]{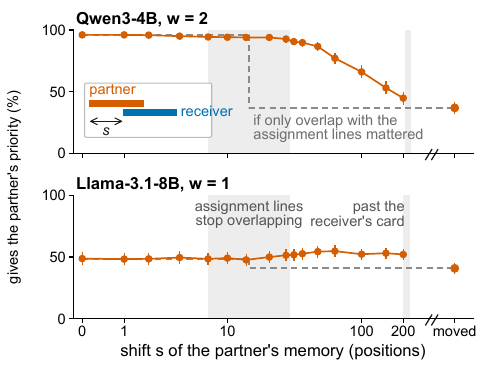}
  \caption{\textbf{Misclaiming as the partner's memory is shifted away from the receiver's.}
  The partner's memory starts $s$ positions before the receiver's (inset); ``moved'' is the full move of \cref{fig:move}. Shading (labelled in the lower panel, the same in both): where the partner's priority stops overlapping the receiver's assignment lines, and where it passes the start of the receiver's card; each band spans the positions for the two line orders. Dashed: misclaiming if it fell to the moved level as soon as the lines stop overlapping. Horizontal axis logarithmic, with the overlapping placement (0) and the full move added. Qwen at $w=2$, Llama at $w=1$; 240 episodes per model, separate from the main study, run at every shift; bars are 95\% CIs.}
  \label{fig:shift}
\end{figure}

The move does two things at once: it ends the overlap between the two memories, and it puts the partner's memory farther from the question than the receiver's. Shifting the partner's memory back only part of the way, by 1 to 200 positions (\cref{fig:shift}), cannot separate these two, because the overlap shrinks as the distance grows. Each card is 264 tokens long in Qwen and 256 in Llama, so even a shift of 200 leaves part of the two cards overlapping. The shifts do test two simpler accounts: that the two memories must line up token for token, and that misclaiming stops as soon as the partner's priority no longer lies alongside the receiver's assignment lines. %

In Qwen, misclaiming declines smoothly as the shift grows (\cref{fig:shift}), from 96.0\% at overlap to 44.8\% at a shift of 200, approaching the 36.9\% of the full move. %
Shifts of one or two positions change misclaiming by less than half a point in either model, so exact alignment does not matter. %
Misclaiming stays high after the partner's priority has moved past the receiver's assignment lines. Averaged over shifts of 48, 64 and 100, Qwen's misclaiming is still 39.7 points above its moved level, far above the dashed curve in \cref{fig:shift}, which marks where misclaiming would fall if overlap with those lines were all that mattered. Varying the order of the two lines, which changes where the priority leaves them, points the same way. %
The decline also begins well before the priority reaches the start of the receiver's card. %

In Llama, misclaiming stays between 47.7\% and 54.8\% at every partial shift, including a shift of 200, where in half of the episodes the partner's priority already lies before the start of the receiver's card. It falls to 40.8\% only after the full move. %
In Llama the order of the two assignment lines on the cards matters even at the overlapping placement: misclaiming is 12.5 points higher when the partner's own line comes first. %
In both models the partner's fact is used in every answer at every shift, and accuracy on the general-knowledge questions varies by less than two points across shifts. Misclaiming therefore needs neither exact alignment nor overlap with the assignment lines: in Qwen it falls gradually as the memories separate, and in Llama only once they are fully apart. %

\section{Further results in Llama-3.1-8B}\label{app:llama}

At double weight Llama's misclaiming falls after the move from 85.1\% to 72.4\%, and the priority it says it is using now from 93.8\% to 19.2\%: the same split as at equal weight. %

Asked ``Which priority do you think Lambda was assigned?'', with the partner's name, the receiver gives the partner's priority less often after the move. At double weight the share falls from 62.3\% to 43.2\% in Qwen and from 79.1\% to 66.9\% in Llama. It falls by more with the third-person card: from 96.0\% to 48.7\% in Qwen and from 96.5\% to 60.0\% in Llama. %
Some of the answers that no longer name the partner's priority now name the receiver's own. In Qwen, others name Robin's priority, which is also what both models mostly answer without the link, when no card states the partner's priority. %
Llama attributes its own priority to the partner even when no partner priority is available at all. With a copy of its own memory linked at double weight and the overlapping placement, it does so in 57.3\% of answers, against 4.3\% without the link and 11.0\% for Qwen in the same condition. %

At equal weight, yes answers to the insertion question fall after the move, from 34.4\% to 17.9\% with the partner's memory and from 29.4\% to 18.5\% with a copy of its own. At double weight they rise, from 39.2\% to 88.2\%, and nearly as much for a copy of its own memory, from 30.8\% to 78.1\%; the AUC of 0.57 is computed at this weight. %

\section{A surcharge stated on both cards}\label{app:conflict}

In these episodes both cards state a surcharge on Thursdays, with different amounts, and everything else is as in the main study (\cref{fig:conflict}). At equal weight the link raises the share of answers giving the partner's surcharge from 0 to 50.4\% in Qwen and misclaiming from 0 to 53.8\%. The move takes back about a sixth of the first rise (8.3 of 50.4 points) but 86\% of the second (46.3 of 53.8). In Llama the link raises the partner's surcharge to 49.9\% and misclaiming to 51.2\%, and the move lowers both by 6.7 points. %

At double weight the pattern is similar. In Qwen the share of answers giving the partner's surcharge falls after the move from 85.3\% to 82.9\% ($-2.3$ points), while misclaiming in the same episodes falls from 96.2\% to 35.8\%. The move thus takes back 3\% of the link's effect on the surcharge and 63\% of its effect on misclaiming. In Llama at double weight the surcharge falls from 58.9\% to 51.9\% ($-7.0$), against $-12.1$ for misclaiming, and the shares are 12\% and 14\%. Without a competitor, when the receiver's card has the filler line instead of a surcharge, the partner's surcharge is used in every answer at both placements in both models. %

At the overlapping placement and equal weight, the receiver's own surcharge has no advantage. Summed over the two ways of assigning each pair of amounts to receiver and partner, the share of answers giving the partner's amount is 1.01 in Qwen and 1.00 in Llama, and 1.08 and 1.02 for the priorities; 1 means that ownership makes no difference. At double weight the sums are higher (Qwen 1.71 for the amounts and 1.92 for the priorities; Llama 1.18 and 1.73), since the partner's memory then weighs more. %

Moving a copy of the receiver's own memory leaves its own surcharge and its own priority in every answer at both placements. In these episodes Llama at double weight loses more general-knowledge accuracy than in the main study (83.5\% and 85.7\% against 95.7\% without the link). In some double-weight arms Qwen's total probability on the answer options of one general-knowledge question is 0.939--0.965. %

\section{The brain-bridge thought experiment}\label{app:bridge}

Brain-to-brain interfaces have so far passed only simple, mostly binary signals between people \citep{rao2014direct,jiang2019brainnet,nam2021direct}, and deeper connections remain proposals. \citet{hirstein2008mindmelding} proposed wiring one person's posterior representations to another person's prefrontal executive processes, and expected the receiver to have the other's content while recognising that it was not its own. Our receiver kept the two apart only when the partner's reflection arrived as a signed text message: it then named the partner's priority as the partner's and never gave it as its own assignment (\cref{sec:phenomenon}). Through the memory link, which like Hirstein's proposal joins internal states directly, it gave the partner's priority as its own original assignment, in about half of its answers at equal weight in both models. Its report of an inserted memory did not separate the partner's memory from a copy of its own (\cref{sec:unsettled}). For this route, our results do not match this expectation. %

A human bridge experiment could face the problem our receiver shows. A person connected by a bridge who behaved like our receiver would report the other's content as their own while describing no second participant, so the first-person evidence the bridge was meant to supply would be misattributed. Such an experiment cannot take the subject's word for whose content is whose. It has to hand out private assignments whose owners are known, measure access, ownership and report separately, and vary the route and the layout of the link. Hirstein's proposal fixes one route. It leaves open the question on which our results turn: whether the arriving representations lie over the receiver's own, like our overlapping memories, or arrive in a place of their own, like our moved memory or a signed message.

\section{Cache layouts of systems that share caches}\label{app:layouts}

\Cref{tab:layouts} lists how published systems place a foreign cache, read from their public code at the commits given, or from their papers. Because rotary attention depends only on the distance between query and key, a cache placed before the receiver's context has the geometry of our moved placement. Different contents given the same positions have the geometry of our overlapping placement. Systems that re-rotate a reused cache give it positions of its own in the new context, as our move does. Cache-to-Cache fuses the two caches rather than concatenating them, so it falls outside both placements. These systems pass caches between different agents, tasks or documents rather than between two copies reading aligned cards, so the table concerns layout, not a replication. Where a system joins caches by plain concatenation, our equal weight is the matching setting: there the prepended layout gives misclaims in 5.0\% of answers in Qwen and 43.4\% in Llama, and the overlapping layout in 49.3\% and 49.1\% (\cref{sec:phenomenon,sec:move}). %

\makeatletter\setlength{\@fptop}{0pt}\makeatother
\begin{table}[t]
  \caption{\textbf{Where published systems place a foreign cache.} Commit prefixes refer to each system's public repository; ``paper'' means the layout is taken from the paper. ``Plain concatenation'' means that the foreign cache joins the receiver's attention with no added weight, which is our equal weight.}
  \label{tab:layouts}
  \centering
  \small
  \begin{tabular}{@{}>{\raggedright\arraybackslash}p{1.45in}>{\raggedright\arraybackslash}p{1.75in}>{\raggedright\arraybackslash}p{2.5in}l@{}}
    \toprule
    Layout & System & Weight on the foreign cache & Read at \\
    \midrule
    Prepended (our moved geometry) & KVComm \citep{shi2026kvcomm} & Plain concatenation, selected layers only & 1fa086e \\
     & LatentMAS \citep{zou2026latentmas} & Plain concatenation (default backend) & 9a9e4d3 \\
    \midrule
    Re-rotated to new positions & KVCOMM \citep{ye2025kvcomm} & No added weight; estimated offsets added to keys and values & 48ca0b3 \\
     & CacheBlend \citep{yao2025cacheblend} & None described; a few tokens recomputed & paper \\
     & Block-Attention \citep{ma2025blockattention} & None described; model fine-tuned & paper \\
    \midrule
    Same content, same positions & Cache-to-Cache \citep{fu2026cache} & Not concatenated: fused into the receiver's cache by a learned module & 5eac72b \\
    \midrule
    Different contents, same positions & Parallel Context Windows \citep{ratner2023pcw} & Plain concatenation & e6d3100 \\
    (our overlapping geometry) & APE \citep{yang2025ape} & Attention to the contexts rescaled by a temperature and a factor & 52513bc \\
     & ParaThinker \citep{wen2025parathinker} & Learned per-path embeddings; model fine-tuned & 6d98fae \\
     & MemOS cache merge \citep{memos2025} & Plain concatenation, not on the default chat path & 64388b5 \\
    \bottomrule
  \end{tabular}
\end{table}

\end{document}